\documentclass[10pt,twocolumn,letterpaper]{article}
\usepackage[toc,page]{appendix}

\usepackage{iccv}
\usepackage{times}
\usepackage{epsfig}
\usepackage{graphicx}
\usepackage{amsmath}
\usepackage{amssymb}
\usepackage{ctable}
\usepackage{tabularx,multirow,booktabs,blindtext}
\usepackage{arydshln}
\usepackage{makecell}
\usepackage[sort,numbers,square,comma]{natbib}
\usepackage[usestackEOL]{stackengine}
\usepackage{soul}
\usepackage[accsupp]{axessibility}

\usepackage[pagebackref=true,breaklinks=true,letterpaper=true,colorlinks,bookmarks=false]{hyperref}

\newcolumntype{Z}{>{\arraybackslash}X}
\newcolumntype{H}[1]{>{\hsize=#1\hsize\centering\arraybackslash}X}
\newcolumntype{C}{>{\hsize=0.8\hsize}Z}
\newcolumntype{D}{>{\hsize=1.2\hsize}Z}
\newcommand*{\crosssymbol}{%
    \text{%
      \raise 1ex\hbox{%
        \rlap{\vrule height.2pt depth.2pt width .75ex}%
        \hbox to .75ex{\hss\vrule height .5ex depth 1ex\hss}%
      }%
    }%
}
\def\iccvPaperID{10235} 

\ificcvfinal\fi

\begin{document}

\title{Towards Discovery and Attribution of Open-world GAN Generated Images}

\author{Sharath Girish\thanks{First two authors contributed equally \newline \hspace*{15pt}To appear at ICCV 2021}\\
{\tt\small sgirish@cs.umd.edu}
\and
Saksham Suri\footnotemark[1]\\
{\tt\small sakshams@cs.umd.edu}
\and
Saketh Rambhatla\\
{\tt\small rssaketh@umd.edu}
\and
Abhinav Shrivastava\\
{\tt\small abhinav@cs.umd.edu}
\and
{University of Maryland, College Park}
}

\maketitle
\ificcvfinal\thispagestyle{empty}\fi

\begin{abstract}
   With the recent progress in Generative Adversarial Networks (GANs), it is imperative for media and visual forensics to develop detectors which can identify and attribute images to the model generating them. Existing works have shown to attribute images to their corresponding GAN sources with high accuracy. However, these works are limited to a closed set scenario, failing to generalize to GANs unseen during train time and are therefore, not scalable with a steady influx of new GANs. We present an iterative algorithm for discovering images generated from previously unseen GANs by exploiting the fact that all GANs leave distinct fingerprints on their generated images. Our algorithm consists of multiple components including network training, out-of-distribution detection, clustering, merge and refine steps. Through extensive experiments, we show that our algorithm discovers unseen GANs with high accuracy and also generalizes to GANs trained on unseen real datasets. We additionally apply our algorithm to attribution and discovery of GANs in an online fashion as well as to the more standard task of real/fake detection. Our experiments demonstrate the effectiveness of our approach to discover new GANs and can be used in an open-world setup. More details will be made available on our~\href{http://www.cs.umd.edu/~sakshams/project_page/gan_wild.html}{project page}.
\end{abstract}
\section{Introduction}

With an ever increasing number of GANs introduced each year, concerns about the malicious use of this technology, especially in the case of social media content \cite{shu2017fake,rossler2018faceforensics,rossler2019faceforensics++} are increasing at an alarming rate which can have adverse impacts on public security and privacy. Therefore, a plethora of works have been proposed which focus on the real/fake detection task \cite{marra2018detection,wang2020cnn,zhang2019detecting}. However, it is also important to focus on the problem of attribution, i.e. identifying the source of these images. Attributing images to their sources can potentially deter malicious organizations and hold them accountable by leading legal proceedings. Additionally, as GANs are becoming part of commercial services such as face animation applications, their popularity draws piracy and plagiarism \cite{yu2019attributing} which is an attack on intellectual property. Therefore, it is pertinent to develop effective techniques to attribute images to specific sources.

\begin{figure}[t]
  \includegraphics[width=\linewidth]{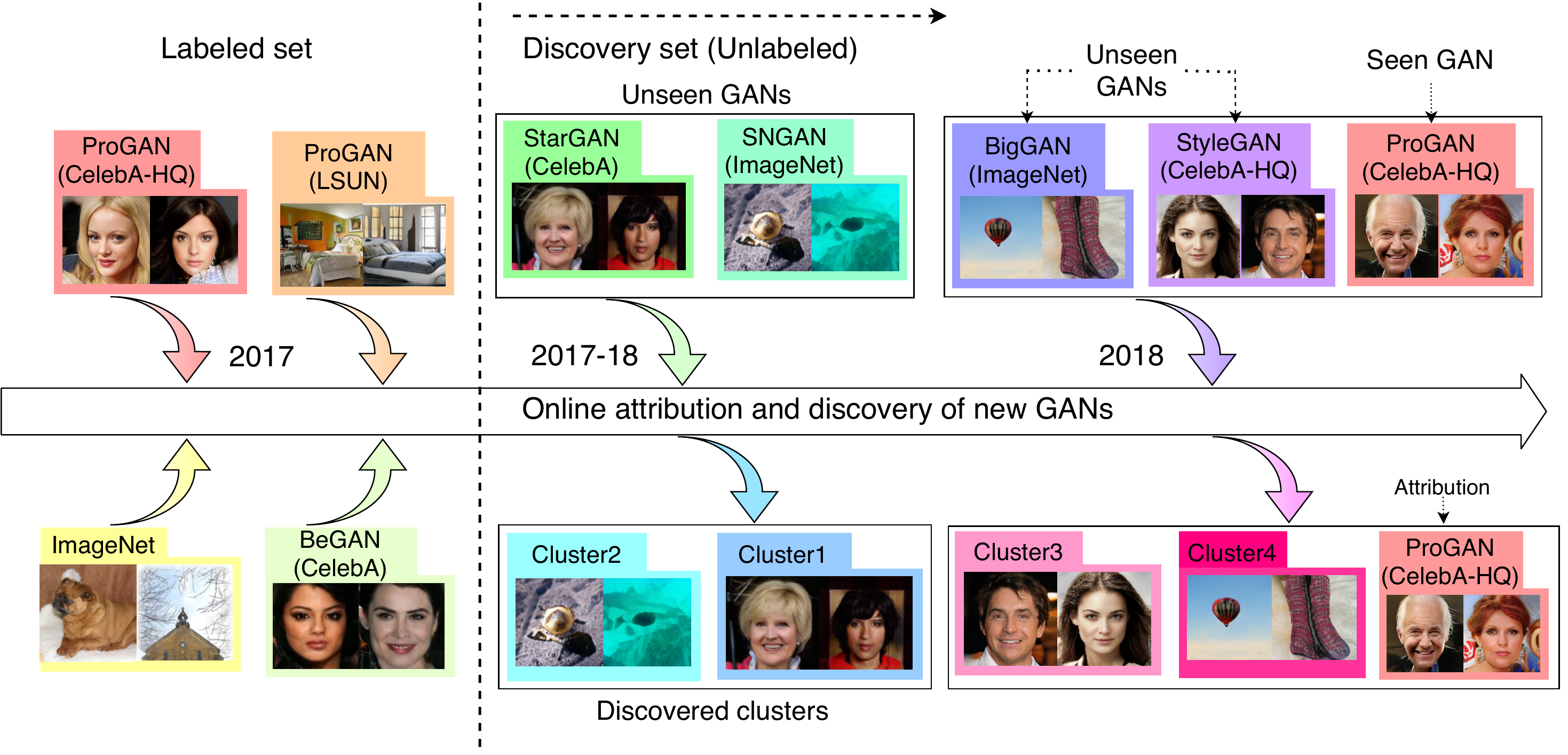}
\vspace{-10pt}
  \caption{A plethora of GANs are released every year, and there could be a set of images that come from several unknown sources. Our approach is capable of discovering and attributing unknown GAN sources while requiring label supervision for only an initial small set of GANs. We attribute with high accuracies, seen GANs from a set of images as well as identify and cluster unknown GAN sources with high purities.}
  \vspace{-0.2in}
\label{fig:teaser}
\end{figure}

To address this problem, \cite{yu2019attributing,8695364,albright2019source} perform attribution for multiple GAN architectures and obtain high classification accuracies. However, they are limited to the closed-world setup as they attribute only to the GANs seen during training and are incapable of identifying unseen GANs. Such a setup is infeasible in practical scenarios where there are a large number of images belonging to sources not seen during training. This raises the question of whether we can discover these new sources and group together the set of images which are generated by them. We term this problem as ``GAN discovery and attribution'' as it involves attributing images to known sources as well as discovering unknown sources. This is a much more challenging and real world setup as the number of sources are unknown and keep increasing. Additionally, there can be a significant domain shift based on the dataset type of GAN generated images. 

Many works such as \cite{wang2020cnn,yu2019attributing, 8695364} show that GANs leave unique artificial signatures in the images they generate. We exploit this information to implicitly identify signatures and cluster images belonging to unseen sources together while also attributing images to seen sources. We propose a novel iterative pipeline which utilizes a fixed set of images, labeled according to their corresponding sources, and perform GAN attribution and discovery on an unlabeled set of images. Our approach generalizes to an open-world setup where images in the unlabeled/discovery set are not restricted to be from the labeled class sources.
Additionally, due to the iterative nature of our pipeline, we can continuously discover images from new GANs added to our discovery set in an online manner. Our approach only requires labels for an initial set of images from real datasets and a few GANs trained on these real datasets with each real/GAN source representing a separate class. While we can discover unseen GANs trained on these real datasets, we additionally show through experiments that we can discover new real datasets and GANs trained on these new datasets as well without them being present in the initial labeled set.

Attribution and discovery in an open-world setup requires us to separate images belonging to seen sources during training from the unseen sources. 
We therefore introduce an explicit out-of-distribution (OOD) step using the deep network features to separate the images belonging to the two types of sources. We propose to incorporate the Winner Take All (WTA) hash~\cite{yagnik2011power} which, to the best of our knowledge, has previously never been used for OOD detection. Additionally, we obtain clusters for the OOD images and perform merge and refine steps to improve the grouping of the unknown GANs using 1-Nearest Neighbour (NN) graphs and kernel SVMs, respectively. We combine these components into a single unified pipeline which is executed iteratively for improving the features and clusters while attributing seen sources and discovering new GAN sources.

Through extensive experiments, we demonstrate the capability of our approach in an open-world setup. We show the efficacy of our approach to generalize to a wide range of dataset setups. We also analyze the importance of the various stages. Additionally, we provide an approach to apply our algorithm for the problem of real/fake image detection and show competitive results on a variety of dataset setups.

We summarize our contributions as follows: \textbf{1)} We introduce a new problem for discovering and attributing images from real and GAN sources in an open-world setup; \textbf{2)} We propose a novel iterative pipeline consisting of several components such as OOD detection, clustering and merge and refine stages providing a strong benchmark for this task, and; \textbf{3)} We analyze the capability of our approach to discover GANs on a variety of dataset setups and also present several insights into the various stages of our pipeline. 

\section{Related Work}

\noindent\textbf{OOD Detection and Open Set Recognition:} 
Several works~\cite{liang2017enhancing, lee2018simple} have tackled OOD detection but require an OOD dataset for tuning hyperparameters, which is not possible as open-world knowledge is not known apriori. \cite{hsu2020generalized} removes this constraint but requires modification of the training setup to decompose confidence scores into two probabilities.

On similar lines is the task of open set recognition \cite{scheirer2012toward}. \cite{jain2014multi,oza2019c2ae,rudd2017extreme} use the Extreme Value Theory to discard unknown samples
but require setting thresholds for reconstruction errors and/or probability values to detect OOD samples which requires careful tuning for each dataset.  \cite{boult2019learning} provides a detailed survey of more works in this area. 

\noindent\textbf{Open World learning:}
While Open Set Recognition only rejects the unseen classes, Open World learning \cite{bendale2015towards} also focuses on reasoning about the unseen classes. \cite{xu2019open} tackles this problem using meta classifiers but are limited to the product classification problem. \cite{hsu2017learning,han2019learning,wang2020open,han2020automatically} also focus on a similar problem but require the unlabeled set to only contain unseen classes and knowledge about number of unseen classes in some cases.

\noindent\textbf{Rank correlation:} 
\cite{yagnik2011power} compute the WTA hash which are ordinal embeddings providing a highly non-linear sparse transformation of the feature vector.
\cite{dean2013fast} use this hashing algorithm for performing fast large scale object detection. To the best of our knowledge, no work utilizes ranking based measures for OOD detection.

\noindent\textbf{Clustering:} Clustering is a highly explored field yet there is no one-size fits all solution.~\cite{sarfraz2019efficient} use a first Nearest Neighbours (1-NN) graph to perform parameter free clustering. Inspired from their work, we use a similar 1-NN graph for our merge step. \cite{yan2020clusterfit} perform K-means clustering on a network's features and retrain the network using the cluster labels as pseudo-labels. Our approach partly involves this setup but contains several other components such as OOD detection and merge and refine steps. Spectral clustering ~\cite{yang2019deep, ng2002spectral, von2007tutorial}  is another common approach 
but requires 
eigenvalues for a large Laplacian which is not tractable for large datasets, as is our case.  Another common direction is training a deep network ~\cite{yang2016joint,xie2016unsupervised,ijcai2017-243,ren2019semi,yang2019learning} which learns embeddings/clusters based on minimizing an objective function.  However, these require careful training so as to not diverge while learning the features in an unsupervised manner.

\noindent\textbf{Real/fake detection and GAN attribution:} 
A plethora of works \cite{rossler2018faceforensics, rossler2019faceforensics++,marra2018detection,wang2020cnn} exist for the problem of real/fake detection but are only limited to this binary classification problem and are not directly applicable to GAN attribution and discovery.
~\cite{wang2020cnn,yu2019attributing,8695364} tackle this problem but are, however, limited to the GANs that they train on and fail to generalize in an open world setup. \cite{marra2019incremental} propose a more dynamic approach to incrementally include GANs for attribution but require clean datasets with images coming from only a single GAN source which does not hold in practice, as images could be generated from multiple sources. To the best of our knowledge, there exists no work dealing with open-world GAN discovery and attribution which is a much harder task than just real/fake detection or closed set GAN attribution.

\begin{figure*}
\centering
\includegraphics[width=\linewidth]{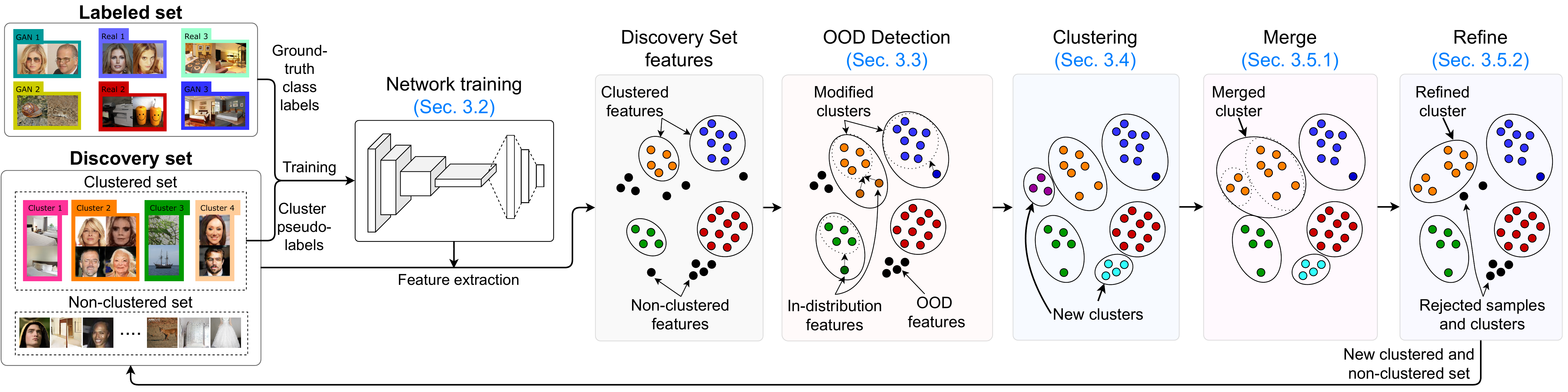}
  \caption{Illustration of our algorithm, where we iteratively discover new classes and retrain our network using them as pseudo-labels.}
  \vspace{-0.08in}
\label{fig:pipeline}
\end{figure*}

\section{Proposed Approach}


\subsection{Overview}
In this section, we briefly describe our approach as shown in Fig. \ref{fig:pipeline}. Our initial labeled set consists of $n_s$ images corresponding to the seen classes and is denoted by $\mathcal{I}_s = \{I_{s_1},I_{s_2},...,I_{s_{n_s}}\}$, and their ground truth class labels, denoted by $\mathcal{Y}_s = \{y_{s_1},y_{s_2},...,y_{s_{n_s}}\}$. The discovery set consists of $n_t$ unlabeled images, from both seen and unseen classes, and is denoted by $\mathcal{I}_t=\{I_{t_1},I_{t_2},...,I_{t_{n_t}}\}$.
Our pipeline proceeds iteratively, and at any point in the pipeline, our discovery set is partitioned into $\mathcal{I}_c$ and $\mathcal{I}_n$.  $\mathcal{I}_c$ is a set of $n_c$ clustered images with predicted labels $\hat{\mathcal{Y}}_c$ while  $\mathcal{I}_n$ is a set of images which could be potentially clustered in future iterations.
In each iteration, we improve the predicted labels in the clustered set ($\mathcal{I}_c$)  and add new samples from the non-clustered set ($\mathcal{I}_n$) into the clustered set. We do this via several stages using algorithms or tools which have previously not been applied for the specific tasks. We also combine the various stages in a unified manner for iteratively improving the features and clusters. 

\smallskip
\noindent\textbf{Network training}: Our network consists of a feature extractor $f(\cdot)$ and classifier $g(\cdot)$. We train the network in a supervised manner using the two sets of images and labels,  labeled set $\left(\mathcal{I}_s,\mathcal{Y}_s\right)$ and clustered set $\big(\mathcal{I}_c,\hat{\mathcal{Y}}_c\big)$. 

\smallskip
\noindent\textbf{Out-of-distribution detection}: We use $f(\cdot)$ to extract features for $\mathcal{I}_c$ and $\mathcal{I}_n$ and perform OOD detection. This stage predicts samples from $\mathcal{I}_n$ to be in-distribution or OOD with respect to the clusters in $\mathcal{I}_c$. The in-distribution samples are classified using the classifier and attributed to $\mathcal{I}_c$ with the corresponding predicted labels. 

\smallskip
\noindent\textbf{Clustering}: We use the K-means algorithm to overcluster the remaining samples in $\mathcal{I}_n$. These clusters are then added to the clustered set $\mathcal{I}_c$ with a new set of labels based on the cluster labels. At the end of this stage all samples have a predicted label and the non-clustered set, $\mathcal{I}_n$, is empty. 

\smallskip
\noindent\textbf{Merge and refine}: To deal with overclustering we perform merge and refine operations. Specifically, coherent clusters are merged to reduce the number of clusters. This reduces the purity of the clusters and hence a refine operation is performed which throws away impure clusters, or samples likely to not have belonged to their existing clusters. The rejected samples are added to the non-clustered set $\mathcal{I}_n$. At the end of this stage, we have a new clustered set, $\mathcal{I}_c$ along with its predicted labels $\hat{\mathcal{Y}_c}$, and non-clustered set, $\mathcal{I}_n$. The four steps described above are then repeated.
We now describe each of the steps enumerated above in detail.




\subsection{Network Training}
\label{ss:step1_classifier}
 This stage involves training the network using the cluster labels $\hat{\mathcal{Y}_c}$ corresponding to $\mathcal{I}_c$ and $\mathcal{Y}_s$ corresponding to $\mathcal{I}_s$ in a supervised manner. The network consists of a feature generation network $f(\cdot)$ parameterized by $\theta_f$, constructed using an off-the-shelf CNN followed by a few fully connected layers to reduce the dimensionality. The classification part of the network $g(\cdot)$ parameterized by $\theta_g$ involves a fully connected layer followed by the softmax function.
 
The parameters of the network $\theta_f,\theta_g$ are optimized as per the following expression:
\vspace{-0.05in}\begin{multline}
    \displaystyle \min_{\theta_f,\theta_g}\Bigg[ \frac{1}{n_c}\sum_{i=1}^{n_c}\mathcal{L}\left(g_{\theta_g} (f_{\theta_f}(I_{c_i})),\hat{y}_{c_i}\right)
    + \\  \frac{1}{n_s}\sum_{j=1}^{n_s}\mathcal{L}\left(g_{\theta_g}(f_{\theta_f}(I_{s_j})),y_{s_j}\right) \Bigg],
    \label{eq:cross_entropy_loss}
\end{multline}
\vspace{-0.05in}where $\mathcal{L}$ is the cross-entropy loss, $I_{c_i}$ and $\hat{y}_{c_i}$ are the $i^\text{th}$ images and labels from $\mathcal{I}_c$ and $\hat{\mathcal{Y}_c}$ respectively while $I_{s_j}$ and $y_{s_j}$ are the $j^\text{th}$ images and labels from $\mathcal{I}_s$ and ${\mathcal{Y}_s}$ respectively. 
Subsequent to network training, we use the feature generation network to extract the features $\mathcal{X}_c$ and $\mathcal{X}_n$ corresponding to the clustered set of images $\mathcal{I}_c$, and non-clustered set of images $\mathcal{I}_n$, respectively.
\subsection{Out-of-distribution detection}
\label{ss:anomaly}
We utilize the WTA hashing algorithm proposed by Yagnik \etal \cite{yagnik2011power} who show that ordinal representations of feature vectors provide strong nonlinear transformations and demonstrate their algorithm's capability on downstream tasks, such as similarity search and classification. They show that such rank correlation measures are robust to noise unlike cosine or Euclidean based distances. Additionally, Euclidean/cosine based distances are highly sensitive to thresholds used for OOD detection which would require careful hyperparameter tuning for different dataset setups. We refer readers to their work or our supplementary material for a detailed explanation of the WTA hash. 

The WTA hash maps a $d$ dimensional feature vector $\boldsymbol{x}$ to a $H$ dimensional vector $\boldsymbol{x}_H$ with elements lying in $[K]$.
Using this hash for each feature vector, we then represent the distance between any two feature vectors $\boldsymbol{x}$ and $\boldsymbol{y}$, as $d(\boldsymbol{x},\boldsymbol{y})$, which is the Hamming distance between their corresponding hashes.
For each class in set $\hat{\mathcal{Y}}_c = \{\hat{y}_i \in [N], \ i \in [n_c]\}$ ($n_c$ is the number of samples in the clustered set, $N$ is number of clusters), we obtain OOD detectors in the following manner: For a cluster with cluster label $j \in [N]$ and for a feature sample $i \in [n_c]$ in the non-clustered set represented by $\boldsymbol{x}_{n_i} \in \mathcal{X}_n$, we compute the distance of $\boldsymbol{x}_{n_i}$ from each sample in the cluster $j$. We then average these sample distances to get the distance of sample $\boldsymbol{x}_{n_i}$ from cluster $j$, \ie, \begin{equation}
    d_j(x_{n_i}) = \frac{1}{N_j}\displaystyle \sum_{k=1, y_{c_k}=j}^{n_c}d(\boldsymbol{x}_{n_i},\boldsymbol{x}_{c_k}),
\end{equation} where $N_j$ represents the number of samples in cluster $j$. The detector then classifies $\boldsymbol{x}_{n_i}$ as an in-distribution sample of class $j$ if $d_j(x_{n_i}) < t_j$ for a threshold $t_j$ for class $j$.
The threshold $t_j$ is computed using the intra cluster distances for each cluster $j$ and setting a high percentile of these distances as the threshold. By doing so, the algorithm learns different thresholds for different clusters and is controlled only by a single percentile scalar which generalizes across different dataset setups.
A test sample, $\boldsymbol{x}_{n_i}$ is classified as an OOD sample to $\mathcal{X}_c$, if all of the detectors for the clusters classify it as OOD. All in-distribution samples are classified using our classifier and their corresponding labels lie in $\hat{\mathcal{Y}_c}$. The samples are subsequently added to $\mathcal{I}_c$.

\subsection{Clustering}
\vspace{-0.05in}
We now overcluster samples remaining in $\mathcal{I}_n$ by running K-Means on the feature set $\mathcal{X}_n$. We form a high number of clusters in order to get clusters with high purity. Once the clusters are obtained, they are added to the clustered set $\mathcal{I}_c$. Their new labels, corresponding to the cluster labels, are added to $\hat{\mathcal{Y}_c}$.
At the end of this stage, no samples remain in the non-clustered set. More importantly, as we generate a large number of clusters, it makes the clustered set highly fragmented. In order to reduce the number of clusters and improve the purity of the clusters we perform a merge and refine step as explained in the following section.

\subsection{Merge and refine}
\label{ss:step1_merge_refine}
\vspace{-0.05in}
Overclustering results in a highly fragmented cluster set which could belong to the same class. To deal with this, a merge step is performed. Anything less than an ideal merge step results in impure clusters. To improve the purity a refine step is also performed. We discuss these in detail below.

\vspace{-0.1in}
\subsubsection{Merge}
\vspace{-0.05in}
We merge clusters in $\mathcal{I}_c$ using a 1-Nearest Neighbour graph. We obtain centroids, $\boldsymbol{u}_j$, for each cluster $j \in [N]$ ($N$ is the number of clusters) by averaging the features of all samples in the cluster. Using the hashing described in Section \ref{ss:anomaly} for each centroid, we define the distance between two centroid feature vectors $\boldsymbol{u}_i$ and $\boldsymbol{u}_j$, $d(\boldsymbol{u}_i,\boldsymbol{u}_j)$, as the Hamming distance between their corresponding hashes ${\boldsymbol{u}_i}_H$ and ${\boldsymbol{u}_j}_H$.

We use the centroid distances between every pair of clusters to create a directed 1-Nearest Neighbour graph with each node representing a cluster centroid. A directed edge is present from one node to another if the latter node is the nearest neighbour centroid of the former node.
Strongly connected components are computed for this graph and each connected component in the graph is considered to be a merged cluster. 
This stage generates a new set of labels, $\hat{\mathcal{Y}_c}$, for the clustered set $\mathcal{I}_c$.

\vspace{-0.1in}
\subsubsection{Refine}
\label{ss:refine}
\vspace{-0.05in}
As the merge step is not ideal, it reduces the average purity of the clusters. In order to increase it, a refine step is performed to remove impure samples from each cluster. As the ground truth labels are unknown, SVM classifiers are leveraged to obtain a proxy measure for purity. 
\cite{malisiewicz-iccv11,shrivastavaSA11} show that weak SVM classifiers can be fit to a single positive instance with the remaining samples as negatives. Therefore, we use this formulation of weak classifiers that can fit to the majority class distribution of a cluster and mark the samples which do not belong to the majority class as negatives.\\
For each cluster $j \in [N]$, an SVM classifier, $Q_j$, is trained in a one-vs-all manner, where the positive samples belong to cluster $j$ while the rest of the samples in the clustered set are negative samples. After training $Q_j$, we use the SVM to predict the labels for samples in cluster $j$ as positive and negative. The samples which are predicted negative are then rejected and added back into the non-clustered set $\mathcal{I}_n$. If the percentage of predicted positive samples by $Q_j$ in cluster $j$ is below a threshold $\epsilon$, the entire cluster is discarded and all the samples are added to $\mathcal{I}_n$.  

Additionally, some refined clusters might have very few samples and the class distribution for training the network in the next iteration could become long tailed. In order to avoid this issue, we threshold clusters based on their sizes and discard those below a size threshold $\tau$ into $\mathcal{I}_n$.

After the refine step we have a new set of clustered images with their corresponding pseudolabels. These are used along with the seen class train data $\mathcal{I}_s$ in order to train the network for the next iteration.

\subsection{Cluster set initialization}
\vspace{-0.05in}
The start of every iteration of our pipeline requires a clustered set $\mathcal{I}_c$ along with the seen labeled set $\mathcal{I}_s$. For the first iteration, as we do not have any pseudolabels for the discovery set $\mathcal{I}_t$, we train our network using only the set $\mathcal{I}_s$ and their corresponding ground truth labels $\mathcal{Y}_s$. Our OOD detection step then determines whether images in $\mathcal{I}_t$ belong to the seen classes $\mathcal{Y}_s$ or not. In-distribution samples are classified and are added to the clustered set $\mathcal{I}_c$ while OOD samples are added to $\mathcal{I}_n$. At the end of this stage, we now have a clustered and non-clustered set for the discovery set images. The rest of the stages of our pipeline, \ie, K-Means Clustering, Merging and Refinement proceed as explained in the previous sections using the initialized $\mathcal{I}_c$ and $\mathcal{I}_n$. The refine step then produces a set of images in the clustered set with their corresponding cluster labels as pseudo-labels which are used to train the network for the next iteration. Additionally, at every iteration $t$, the feature extractor is initialized with the weights of the previous iteration $t-1$. The classifier is replaced with a new linear layer with weights randomly initialized as number of classes, which is dependent on number of clusters $N$, change across iterations.
The algorithm then proceeds for a few iterations until fraction of undiscovered samples fall below a small threshold. 

\section{Experiments}
\label{sec:experiments}
We now evaluate our approach on real world dataset setups while providing detailed analysis of the several components of our pipeline. In Section \ref{ss:exp_details}, we describe the implementation details. Our labeled dataset consists of images from 4 real datasets as well as from certain GANs trained on these real datasets as shown in Table \ref{tab:default_set}. Together, they make up 12 classes in the labeled set. Our discovery set consists of additional images from these 12 classes as well as from 8 unseen GANs as shown in Table~\ref{tab:default_set} making up a total of 20 classes. We use, by default, this dataset for all our experiments unless mentioned otherwise. Note that the same GAN trained on different datasets corresponds to different classes. 
 Section \ref{ss:baselines} shows extensive comparisons with other related works on GAN attribution and real/fake image detection. Section \ref{ss:analysis} provides several insights into our algorithm and also analyzes several components of our pipeline. Subsequently, we examine the results of our pipeline on varying dataset setups.  Section \ref{sss:num_gans} shows an analysis of number of GANs needed in our labeled set to reliably discover new GANs in the discovery set. Section \ref{ss:new_dataset} changes number of unseen real datasets as well as corresponding GANs in the discovery set and shows the effectiveness of our approach to discover these new classes.

\begin{table}
\centering
\footnotesize
\renewcommand{\arraystretch}{1}
\renewcommand{\tabcolsep}{6pt}
    \centering
    \caption{List of GANs trained on the corresponding 4 real datasets used in our labeled and discovery set. Note that the same GAN can be trained on multiple datasets.}
    \begin{tabular}{@{}lll@{}}
    \toprule
     \textbf{Dataset} & \textbf{Labeled GANs} &  \textbf{Discovery GANs} \\
    \midrule
    \makecell[tl]{CelebA\cite{liu2015faceattributes}}& \makecell[tl]{StarGAN\cite{choi2018stargan},\\ AttGAN\cite{he2019attgan}}& \makecell[tl]{StarGAN, BEGAN\cite{berthelot2017began}, \\ProGAN\cite{karras2017progressive}, SNGAN \cite{miyato2018spectral}, \\AttGAN, MMDGAN\cite{li2017mmd}, \\CramerGAN\cite{bellemare2017cramer}}\\ \hdashline
    
    \makecell[tl]{CelebA-HQ\\\cite{karras2017progressive}} & \makecell[tl]{ProGAN, \\StyleGAN\cite{karras2019style}} & \makecell[tl]{ProGAN, StyleGAN, \\ResNet19\cite{kurach2019large}}\\\hdashline
    ImageNet~\cite{deng2009imagenet} & \makecell[tl]{BigGAN\cite{brock2018large}, \\S3GAN\cite{lucic2019high}}& BigGAN, S3GAN, SNGAN \\ \hdashline
    \makecell[tl]{LSUN\\Bedroom~\cite{yu2015lsun}}& \makecell[tl]{ProGAN, \\MMDGAN} & \makecell[tl]{ProGAN, MMDGAN,\\ SNGAN}\\
    \bottomrule
    \end{tabular}
    \vspace{-0.2in}
    \label{tab:default_set}
\end{table}

\subsection{Experimental details}
 \label{ss:exp_details}
 \vspace{-0.05in}
For our feature extractor, we use the standard ResNet-50 ~\cite{he2016deep} backbone. We add 3 fully-connected layers to reduce the dimensionality of the feature vector to $128$. Another fully connected layer is used as the classification head on top of the feature extractor. The full network is trained in a supervised manner and using cross entropy loss. Every image is resized and center cropped to $256\times256$ except when specified otherwise. We use a batch size of $256$ for our training for each iteration of the pipeline. The weights are optimized using the Adam optimizer with $\beta_1=0.9$, $\beta_2=0.999$ and a fixed learning rate of $0.0001$ throughout our training. For the first iteration, we train our network for $50$ epochs, while for subsequent iterations we train for $100$ epochs, as the network takes longer to converge with additionally discovered samples with noisy pseudo labels. For our OOD detection step using WTA hash described in Section \ref{ss:anomaly}, we use $H=2048$ hashes and a window size of $K=2$. Our clustering stage uses the K-Means algorithm for $500$ clusters initialized using K-Means++ ~\cite{arthur2006k}. For the refine stage, we train SVMs with the RBF-kernel. We set the threshold, $\epsilon=0.5$, for dropping a cluster, as described in Section \ref{ss:refine}. To avoid training on clusters with very few samples, we discard clusters with less than $100$ members.

\textbf{Metrics and analysis:} We evaluate our pipeline on 2 clustering metrics. We use Average Purity as a metric for evaluating the overall purity of our clusters with respect to the true labels of the discovery set. We also use Normalized Mutual Information (NMI), which is another commonly used clustering metric. At various stages or iterations of our pipeline, a small fraction of the discovery set samples remain non-clustered and in order to provide a fair evaluation across different experiments/baselines we attribute all the non-clustered samples to their nearest clusters and evaluate on the full discovery set, unless mentioned otherwise.

\begin{table}
    \centering
    \footnotesize
    \renewcommand{\arraystretch}{1.1}
    \renewcommand{\tabcolsep}{3pt} 
    \setlength{\cmidrulewidth}{0.01em}
\caption{Comparison of our method with baselines derived from ~\cite{yu2019attributing,wang2020cnn, han2020automatically}. We try two fixed setups for number of clusters $k=20,500$ and finally let our approach discover the suitable number of clusters $k=209$. Compared to the 2 baselines, we obtain the highest Average Purity and NMI when number of clusters $k=209$. Ours [only $\S$3.2] corresponds to a single iteration of network training and clustering.  The fully supervised setup is the upper bound when all classes are seen.}
\vspace{0.03in}
\resizebox{\linewidth}{!}{
\begin{tabular}{@{}lcccccc@{}}
\toprule
\multirow{2}{*}{Method} & \multicolumn{2}{c}{$k=20$} & \multicolumn{2}{c}{$k=500$} &\multicolumn{2}{c}{$k=209$}\\
\cmidrule[\cmidrulewidth](l){2-3}
\cmidrule[\cmidrulewidth](l){4-5}
\cmidrule[\cmidrulewidth](l){6-7}
& Avg. Purity & NMI & Avg. Purity & NMI  & Avg. Purity & NMI \\
\midrule
Yu \etal ~\cite{yu2019attributing} & 0.656 &  0.706 & 0.759 & 0.518  & 0.734 &0.554 \\
{Han \etal ~\cite{han2020automatically}}  & 0.680 & 0.709 & - & -  & - & -\\
{Wang \etal ~\cite{wang2020cnn}}  & 0.710 & 0.759 & 0.857 & 0.575 &0.840 & 0.624\\
Ours [only $\S$3.2] & 0.661 & 0.743  & 0.814 & 0.561    & 0.795 & 0.609\\

Ours  &- &- & - &-  & \textbf{0.861} & \textbf{0.724} \\ \hdashline 
Fully supervised & 0.928 & 0.929 & 0.996 & 0.658  & 0.997 &0.728\\
\bottomrule
\end{tabular}
}

\vspace{-0.1in}
\label{tab:baselines}
\end{table}

\subsection{Benchmark Evaluation}
\label{ss:baselines}
\vspace{-0.05in}
As there exists no prior work dealing with open-world GAN discovery, we provide baselines by modifying recent works involving GAN attribution~\cite{yu2019attributing} and real/fake image detection~\cite{wang2020cnn}. We additionally include the recent approach of~\cite{han2020automatically} which deals with novel category discovery.

Yu~\etal~\cite{yu2019attributing} deals with GAN attribution in a closed-world setup and hence cannot be directly incorporated to our problem setup. Therefore, we train their network on our labeled set and obtain features for our discovery set. We cluster the features using K-Means for 3 different values of $k$. $k=20$ corresponds to the true number of classes in our test set while $k=500$ corresponds to an overclustered regime. $k=209$ represents the number of clusters our algorithm returns at the end of 4 iterations. We compare across multiple values of $k$ as Average Purity and NMI are known to be sensitive to number of clusters.

Wang~\etal~\cite{wang2020cnn} tackles real/fake detection and again cannot be directly used in our problem setup. Therefore, we modify their classification head to be multiclass and train their network on our labeled set using their training and preprocessing strategies and extract the features for our discovery set. We provide three similar baselines by performing clustering similar to the baselines generated from ~\cite{yu2019attributing}.

Han~\etal~\cite{han2020automatically} discover novel visual categories but require the discovery set to only contain unseen classes. We therefore use our anomaly detection approach on their features to separate out the seen and unseen classes whose cluster assignments are then predicted separately using their approach. As they require knowledge of number of unseen classes for their predictions, we compare with the $k=20$ setup which corresponds to the true number of classes.

Finally, we provide a baseline for our approach by performing network training and clustering the feature space into $k=20,500,209$ clusters. We also provide an upper-bound for our approach using a fully supervised case where the labeled set consists of images from all classes in the discovery set and perform clustering on the generated features.

The results for these comparisons are provided in Table~\ref{tab:baselines}. Our algorithm achieves the highest Average Purity and NMI compared to all other baselines for the case of $k=209$. For $k=20,500$, \cite{wang2020cnn} outperforms a single iteration of network training and clustering because it does not involve OOD detection, merge or refine for this comparison. However, at the end of 4 iterations, for the case of $k=209$, we significantly outperform all baselines in terms of both Average Purity and NMI. The fully supervised approach provides an upper bound for all 3 cases. Note that we do not compare across number of clusters as Average Purity increases in general with more clusters while NMI decreases.


\begin{table}
    \centering
    \footnotesize
    \renewcommand{\arraystretch}{1.2}
    \renewcommand{\tabcolsep}{6pt}
    \caption{We analyze the effect of the various stages of our pipeline. The number of clusters in the merge step decreases with negligible drop in Avg. Purity and increased NMI. The Refine step further increases the NMI and Avg. Purity by a big margin for the discovered samples. Note that the numbers corresponding to all samples in the refine step are included for the sake of fair comparison but are not actually computed by our approach.}
    \begin{tabular}{@{}lccc@{}}
    \toprule
    Stage &No. of clusters& Avg. Purity& NMI\\
    \midrule
    Clustering & 512 & 0.793 & 0.682\\
    Merge & 391 & 0.792 & 0.689\\
    Refine (Discovered) &111&\textbf{0.849}&\textbf{0.838}\\
    Refine (All) &111&0.772&0.720\\
    \bottomrule
    \end{tabular}
    \vspace{-0.1in}
    \label{tab:stages}
\end{table}

\begin{table}
    \centering
    \footnotesize
    \renewcommand{\arraystretch}{1.1}
    \renewcommand{\tabcolsep}{6pt}
    \caption{We evaluate our algorithm over multiple iterations. Avg. Purity, NMI and \% of discovered samples progressively increases.}
    \begin{tabular}{@{}ccccc@{}}
    Iteration & \makecell{Avg.\\ Purity}& NMI &\makecell{\% Samples\\ Clustered} & \makecell{Sources\\Discovered}\\
    \midrule
    1& 0.772 &0.720&72.5&16/20\\
    2& 0.853 &0.724&88.8&20/20\\
    3& 0.861 &0.724&92.3&20/20\\
    4& \textbf{0.861} & \textbf{0.724}&\textbf{93.7}&\textbf{20/20}\\
    \bottomrule
    \end{tabular}
    \vspace{-0.15in}
    \label{tab:iteration}
\end{table}

\vspace{-3pt}
\subsection{Ablation Study}
\label{ss:analysis} 
\vspace{-0.05in}
Our algorithm is fairly robust to the various hyperparameter values used in our stages. Experiments for varying hyperparameter values are shown in the supplementary material. In this section, we analyze the importance of each stage and the progress of our pipeline over multiple iterations.

We evaluate the effect of Clustering, Merge, and Refine  stages in the first iteration of our pipeline. The results are summarized in Table \ref{tab:stages}. Note that the Average Purity drops only slightly in the merge step while the number of clusters drop significantly demonstrating the effectiveness of the 1-NN merge step explained in Section  \ref{ss:step1_merge_refine}. From the merge to the refine step, Average Purity drops for the full discovery set as many samples remain undiscovered and we evaluate the metric over the full discovery set by na\"ively attributing them to the nearest cluster. However, the metrics evaluated on only the discovered samples increase significantly which shows that SVMs can identify the pure clusters and samples while rejecting the impure ones. 

Next, we evaluate our pipeline over multiple iterations. We show the results in Table \ref{tab:iteration}. The pipeline discovers only a small fraction of images and GANs in the first iteration while in subsequent iterations, more samples are added to the clustered set and more GANs are discovered. Average Purity and NMI both increase or remain constant over the four iterations which shows the effectiveness of our approach to discover as well as improve clusters. Our OOD stage obtains an accuracy of $86.97\%$ for seen classes, $99.87\%$ for unseen classes and $92.87\%$ overall. The high unseen class accuracy is because of setting a lower threshold to reduce false negative errors which do not get corrected in subsequent stages.
\begin{figure*}
\centering
  \includegraphics[width=\linewidth]{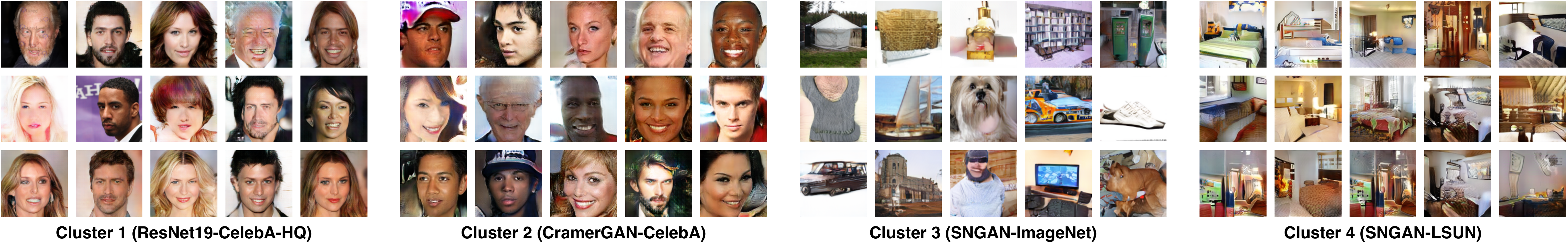}
  \caption{Samples from clusters discovered by our approach for unseen GANs with the majority class in parenthesis. It can be noticed that they are not just focusing on the object structure and semantics rather the underlying source.}
\label{fig:cluster_vis}
\vspace{-0.15in}
\end{figure*}


\vspace{-0.03in}
\subsection{Varying dataset setups}
\vspace{-0.05in}
In this section, we provide an analysis by varying the dataset setups based on number of GANs per real dataset in the labeled set or on adding new real datasets and GANs trained on them in the unlabeled set.



\vspace{-0.03in}
\subsubsection{Effect of number of GANs per dataset}
\label{sss:num_gans}
\vspace{-0.05in}
We answer the question of how many GANs per dataset are needed in our labeled set to reliably discover new ones in our discovery set. We have 3 labeled dataset setups: \textbf{1)} Our first setup consists of 4 real datasets: CelebA, CelebA-HQ, ImageNet and LSUN-Bedroom with no GANs;
\textbf{2)} In addition to the 4 datasets in the first setup, our second setup has 4 GANs: StarGAN, ProGAN, BigGAN and MMDGAN trained on the respective datasets; \textbf{3)} In addition to the previous setup, we have 4 more GANs per dataset: AttGAN, StyleGAN, S3GAN and ProGAN. 

In order to fairly evaluate the 3 setups, we use a common discovery set consisting of all the classes in the second setup. Additionally, we have a set of GANs not present in all 3 labeled sets, namely, BEGAN, ResNet19 (from CompareGAN \cite{kurach2019large}), SNGAN and CramerGAN corresponding to the 4 real datasets.
The results are summarized in Table ~\ref{tab:ngan}. Due to most information being present in the labeled set, the third setup performs best on both Average Purity and NMI. Despite the second setup having only a single GAN per dataset, it performs fairly well on the two metrics. On the other hand, the first setup, which does not have any GANs in the labeled set, fails to discover new ones as it cannot see any GAN-related artifacts in the labeled set and thus fails to discriminate based on this during discovery.

\begin{table}[t]
    \centering
    \footnotesize
    \renewcommand{\arraystretch}{1.1}
    \renewcommand{\tabcolsep}{6pt}
    \caption{Varying number of GANs per dataset. We obtain the best metrics with the maximum number of GANs per dataset although discovering fewer samples compared to the first setup.}
    \begin{tabular}{@{}ccccc@{}}
    \toprule
    \makecell{\# of\\GANs} & \makecell{Avg.\\ Purity}& NMI &\makecell{\% Samples\\ Clustered} & \makecell{Sources\\Discovered}\\
    \midrule
    0 &0.497&0.559&\textbf{99.78}&8/12\\
    1 & 0.897 & 0.772 & 94.48&11/12\\
    2 & \textbf{0.954} & \textbf{0.789} & 95.98&\textbf{11/12}\\
    \bottomrule
    \end{tabular}
    \label{tab:ngan}
    \vspace{-0.15in}
\end{table}

\begin{table}
\centering
\footnotesize
\renewcommand{\arraystretch}{1.1}
\renewcommand{\tabcolsep}{4pt}
    \centering
    \caption{Effect of adding new datasets and GANs trained on new datasets at test time. (*) provides the corresponding comparison when the real datasets are present in the labeled set (Sec.~\ref{ss:new_dataset})}
    \resizebox{\linewidth}{!}{
    \begin{tabular}{@{}lcccc@{}}
    \toprule
    Test Set & Purity & NMI & \makecell{Sources\\ Discovered} & \makecell{\# of\\ Clusters} \\
    \midrule
    New Real &0.942  & 0.813& 14/16  & 103\\
    New Real* & \textbf{0.989}& \textbf{0.989}&\textbf{15/16} &56\\
    \midrule
    New GANs & \textbf{0.976} & 0.828 & \textbf{16/16} & 105\\
    New GANs* & 0.95 & \textbf{0.835} & 15/16 & 87\\
    \midrule
    New Real + New GANs& 0.850 & 0.730  & \textbf{20/20} & 141\\
    New Real + New GANs* & \textbf{0.977} & \textbf{0.856}  & 19/20 & 128\\
    \bottomrule
    \end{tabular}
    }
    \vspace{-0.2in}
    \label{tab:new_dat}
\end{table}

\vspace{-0.03in}
\subsubsection{Discovering new dataset images}
\label{ss:new_dataset}
\vspace{-0.05in}
In an open-world setting, the discovery set may contain images from new real datasets not seen in the labeled set along with GAN generated images corresponding to these datasets. To see whether the proposed approach can handle these situations we perform experiments covering 3 setups. Each setup uses the default labeled set in Table \ref{tab:default_set} but additional classes in the discovery set as follows: \textbf{1)} New real datasets: New real datasets namely DTD ~\cite{cimpoi14describing}, FashionGen~\cite{rostamzadeh2018fashion}, and Night and Shoes datasets (from Pix2Pix~\cite{isola2017image}); \textbf{2)} GANs on new real classes: New GANs trained on the four new real world datasets, namely, ProGAN on DTD, DCGAN~\cite{radford2015unsupervised} on FashionGen, and a separate Pix2Pix on Night and Shoes datasets;
\textbf{3)} New Real + New GANs: A combination of GANs and real datasets from the previous two setups.

In order to provide a benchmark for comparison, we show the performance when the four real datasets are in the labeled set (marked with a *). The results are shown in Table \ref{tab:new_dat}. In the first setup the goal is to discover new dataset sources. Our approach discovers most of the sources with high Purity and NMI, although it's performance is lower than the benchmark as expected because the labeled set for the benchmark contains all the classes present in the discovery set.
In the second setup, our method discovers all unseen GANs even though they are trained on unseen datasets unlike the benchmark which does slightly worse in terms of Avg. Purity and number of GANs discovered likely because of the reduced number of final clusters. The third setup is more challenging due to the addition of both unseen datasets and GANs trained on them to the discovery set. However our approach discovers all unseen sources with reliable Average Purity and NMI while its corresponding benchmark does not discover all sources possibly because it restricts itself to lesser but purer clusters with higher NMI.


\vspace{-0.03in}
\subsubsection{Online discovery}
\vspace{-0.05in}
Here we extend our approach to an online setup where new GANs are added to the discovery set in an online fashion based on the chronological order they were published. Our setup consists of 9 GANs from 4 real sources in our labeled set and 4 new GANs in the discovery set. We additionally introduce 2 sets of 3 GANs each in an online fashion. Details of the datasets are provided in supplementary material. We show our results in Table \ref{tab:online}. We train our setup for 2 iterations with the initial discovery set of 17 sources. It can be seen that Average Purity increases in the second step and it also discovers an additional GAN source. When new GANs are introduced in iterations 3 and 5, the performance drops as the network is not trained on the new classes. However, after a single iteration the Average Purity increases significantly and NMI drops only slightly even though number of clusters increase. At the end of 6 iterations, we discover all the GAN sources added on the fly, except one. This shows that our approach works in an online setting, continuously discovering new GANs iteratively.

\begin{table}
\centering
\footnotesize
\renewcommand{\arraystretch}{1.2}
\renewcommand{\tabcolsep}{6pt}
    \centering
    \caption{Evaluation of our algorithm in an online setup. We have 17 sources in our initial discovery set and add 3 sources each at iteration 3 and 5 causing an initial drop in results. The pipeline eventually performs better after training on the new samples.}
    \begin{tabular}{@{}ccccc@{}}
    \toprule
    Iteration & \makecell{Avg.\\ Purity}& NMI & \makecell{\% Samples\\ Clustered} & \makecell{Sources\\ Discovered}\\
    \midrule
    1& 0.846 &0.826&89.39&15/17\\
    2& 0.916 &0.798&92.42&16/17\\
    \hdashline
    3& 0.805 &0.771&95.74&18/20\\
    4& 0.805 &0.744&96.87&19/20\\
    \hdashline
    5& 0.731 &0.716&95.68&22/23\\
    6& 0.802 &0.705&95.36&22/23\\
    \bottomrule
    \end{tabular}
    \vspace{-0.2in}
    \label{tab:online}
\end{table}

\subsection{Real/Fake detection}
\vspace{-0.05in}
We now apply our method to the common problem of real/fake detection. We use the binary classification model from ~\cite{wang2020cnn}, but trained on our labeled set and use majority voting to mark a cluster and all its constituent images as real or fake. We compare this with using the model directly on all samples and compare the performance in Table ~\ref{tab:real_fake} for our original setup and for the three setups defined in Sec. \ref{ss:new_dataset}. We observe that in most settings, we outperform the standard predictions which are evaluated sample-wise. We attribute it to the fact that the clustering is able to correct model's mistakes as it groups samples according to the source. As cluster assignments are less accurate due to increased difficulty of the final setup, our performance is lower but nevertheless, competitive with \cite{wang2020cnn}.

\subsection{Qualitative analysis of clusters}
\vspace{-0.05in}
We visually inspect a few clusters generated by our method to see whether they focus on the semantic information or the GAN source. To this end we visualize random images from some of the highly pure clusters corresponding to unseen GANs trained on ImageNet, LSUN-Bedroom, CelebA and CelebA-HQ. As evident from Fig.~\ref{fig:cluster_vis} the cluster in the case of SNGAN-ImageNet does not seem to be object-specific, while the cluster for SNGAN-LSUN does not focus on specific room decor, lighting conditions, layout etc. Similarly, clusters corresponding to the face datasets seem to be focusing on the GAN source rather than specific facial attributes like expression, orientation, age etc. 
In addition to visualizing these clusters, we also add a qualitative analysis of the merge step in the supplementary material showing sub-clusters that are merged by our pipeline.

\begin{table}
    \centering
    \footnotesize
    \renewcommand{\arraystretch}{1}
    \renewcommand{\tabcolsep}{6pt}
    \caption{We evaluate the real/fake detection accuracy (\%) using the clustering obtained from our network.}
    \begin{tabular}{@{}lcccc@{}}
    \toprule
    Approach & Original & New Real & New GANs & \makecell{New Real +\\New GANs}\\
    \midrule
    Wang \etal ~\cite{wang2020cnn} & 92.56\% & 87.35\% &98.42\%&\textbf{89.09\%}\\
    Ours & \textbf{98.62\%}&\textbf{89.84\%}&\textbf{99.10\%}&83.33\%\\
    \bottomrule
    \end{tabular}
    \vspace{-0.2in}
    \label{tab:real_fake}
\end{table}

\vspace{-0.08in}
\section{Conclusion}
\vspace{-0.05in}
We proposed a new problem of open-world GAN discovery and attribution. We presented an iterative approach to discover and attribute images from multiple GAN sources in a discovery set. Our framework discovers and groups GANs not seen during training by implicitly focusing on GAN-based fingerprints. We show ablation studies for the different components of our pipeline. We also show the generalization of our approach to various dataset setups and its extension to an online setting. As there have been no works addressing this problem, we compare with several baselines based on state-of-the-art related works and provide a strong benchmark for this task. Even though our approach works in an online setup, network training is an expensive step for each iteration. One potential direction for future work is to utilize approaches from continual learning literature~\cite{li2017learning} for faster training, to learn in a never-ending setup discovering new GANs on-the-fly. We hope, given the general formulation of the stages, our framework is utilized for other similar tasks as well. To facilitate such exploration of different scenarios we plan to release the toolset we have developed for our work to bolster future research in this area. 

\smallskip
{\small \noindent\textbf{Acknowledgements.} This project was partially funded by DARPA SemaFor (HR001119S0085), DARPA SAIL-ON (W911NF2020009), and an independent gift from Facebook AI.}


{\small
\bibliographystyle{ieee_fullname}
\bibliography{egbib}
}

\appendix
\appendixpage

\section{Additional experimental details}
For our feature extractor, we use 3 fully-connected layers. Each fully connected layer is followed by the ReLU activation unit and Dropout with a drop probability of $0.5$ during train phase for regularization. The first layer maps the input $2048$ dimensional vector to $512$ dimension. The second layer maintains the number of activation units at $512$ while the third one downsamples it to $128$ which is the final dimension of the feature vector we use for subsequent stages. For all our experiments in the supplementary, we train on $128\times128$ sized images. We set a percentile threshold of $0.9$ for our out-of-distribution detection stage. Our cluster merging algorithm using the 1-Nearest Neighbour Graph is a 2 staged setup which initially merges the newly obtained clusters from K-Means in the previous stage and then merges the entire clustered set. We adopt this 2 staged setup as K-Means overclusters the discovery set and requires merging before merging with the clusters in the clustered set. For training SVMs, we use the GPU-accelerated library of ThunderSVM~\cite{wenthundersvm18}. 
\section{Additional dataset details}
Table~\ref{tab:data} summarizes the class-wise train and test splits used across our experiments. We use a variety of images for multiple image sources to more closely simulate a real world setup. Note that some train images are not used depending on the dataset setup where the image sources could only belong in the discovery set. \\
Our online dataset setup defined in Section. 4.4.3 of the paper consists of GANs in the chronological order they were published or introduced. We have an initial labeled and discovery set as defined in Table \ref{tab:online_set}. After running our pipeline for 2 iterations on this set, we add 3 more GANs to the discovery set: BigGAN and SSGAN \cite{chen2019self}, both trained on ImageNet, StyleGAN trained on CelebA-HQ. We run our pipeline for 2 more iterations on the new discovery set, and add 3 more GANs: ResNet19 and StarGAN-v2 \cite{choi2020stargan}, both trained on CelebA-HQ, S3GAN trained on ImageNet. This is followed by 2 more iterations of network training resulting in a total of 6 iterations for the full online setup. The numbers are as reported in Section 4.4.3 of the main paper.

\begin{table}
\renewcommand{\arraystretch}{1.1}
\renewcommand{\tabcolsep}{6pt}
    \centering
    \small
    \caption{List of GANs trained on the corresponding 4 real datasets used in our labeled and discovery set. Note that the same GAN can be trained on multiple datasets.}
\begin{tabular}{@{}llrr@{}}
\\ \hline
Dataset& Image Source& \makecell[r]{\# of Images\\(Train)} & \makecell[r]{\# of Images\\ (Test)} \\ \toprule
\multirow{8}{*}{CelebA}&Real&$20k$&$10k$\\
&StarGAN&$20k$&$5k$\\
&AttGAN&$20k$&$10k$\\
&BEGAN&$20k$&$10k$\\
&ProGAN&$20k$&$10k$\\
&SNGAN&$20k$&$10k$\\
&MMDGAN&$20k$&$5k$\\
&CramerGAN&$20k$&$10k$\\
\midrule
\multirow{4}{*}{CelebA-HQ}&Real&$20k$&$10k$\\
&ProGAN&$20k$&$10k$\\
&StyleGAN&$20k$&$5k$\\
&ResNet19&$20k$&$10k$\\
\midrule
\multirow{4}{*}{ImageNet}&Real&$20k$&$10k$\\
&BigGAN&$20k$&$5k$\\
&S3GAN&$20k$&$10k$\\
&SNGAN&$15k$&$10k$\\
\midrule
\multirow{5}{*}{LSUN-Bedroom}&Real&$20k$&$5k$\\
&ProGAN&$20k$&$10k$\\
&MMDGAN&$20k$&$5k$\\
&SNGAN&$20k$&$3k$\\
&CramerGAN&$20k$&$10k$\\
\midrule
\multirow{2}{*}{DTD}&Real&-&$10k$\\
&ProGAN&$20k$&$5k$\\
\midrule
\multirow{2}{*}{FashionGen}&Real&$20k$&$10k$\\
&DCGAN&$20k$&$5k$\\
\midrule
\multirow{2}{*}{Night}&Real&$15k$&$5k$\\
&Pix2Pix&$15k$&$10k$\\
\midrule
\multirow{2}{*}{Shoes}&Real&$20k$&$3k$\\
&Pix2Pix&$20k$&$10k$\\

\bottomrule
\end{tabular}
\label{tab:data}
\end{table}

\begin{table}
\renewcommand{\arraystretch}{1.1}
\renewcommand{\tabcolsep}{6pt}
    \centering
    \small
    \caption{Initial labeled and discovery set for our online setup.}
    \vspace{5pt}
    \begin{tabular}{@{}lll@{}}
    \toprule
     \textbf{Dataset} & \textbf{Labeled GANs} &  \textbf{Discovery GANs} \\
    \midrule
    \makecell[tl]{CelebA}& \makecell[tl]{BEGAN, \\MMDGAN,\\
    CramerGAN,\\ProGAN}& \makecell[tl]{BEGAN, MMDGAN,\\
    CramerGAN, ProGAN,\\
    StarGAN, AttGAN,\\
    SNGAN}\\ \hdashline
    
    \makecell[tl]{CelebA-HQ} & \makecell[tl]{ProGAN} &-\\\hdashline
    ImageNet& \makecell[tl]{SNGAN}& - \\ \hdashline
    \makecell[tl]{LSUN\\Bedroom}& \makecell[tl]{ProGAN, \\MMDGAN,\\CramerGAN} & \makecell[tl]{ProGAN, MMDGAN,\\ CramerGAN,SNGAN}\\
    \bottomrule
    \end{tabular}
    \vspace{-5pt}
    \label{tab:online_set}
\end{table}

\section{Additional baseline comparisons}
Section 4.2 of the paper provides comparisons with baselines derived from the works of \cite{yu2019attributing} and \cite{wang2020cnn} by training their methods on our dataset in a multiclass manner. We additionally provide baselines by using features from the pretrained models provided by them which were trained on their datasets. We provide results by performing K-means clustering on the features for $k=20$ and $k=500$ similar to the baselines derived in Section 4.2. The results are shown in Table \ref{tab:additional_baselines} (denoted by *). It can be seen that the features don't generalize across datasets and does worse than the baselines reported in the paper on both the metrics of Average Purity and NMI. \\
Also, as \cite{wang2020cnn} primarily deals with only real-fake classification, we train their method on our dataset but only on the binary real-fake classification task and extract their features. We show the results based on clustering the features for $k=20$ and $k=500$ and reporting results in Table ~\ref{tab:additional_baselines} (denoted by $\crosssymbol$). We see that features generated from the binary classification problem do worse than the multiclass case. This is because the binary classification problem only discriminates between real and fake image sources while grouping the different fake image sources together. This causes less discrimination between the fake image sources harming the clustering performance. We also provide a baseline (denoted by \#) using our approach but adding JPEG and blur augmentations as used by \cite{wang2020cnn}. We see that this degrades the performance compared to our original approach likely because these augmentations destroy valuable high frequency information used for discriminating between GAN sources. Since the baseline performance was lower in these evaluations we did not include them in the main paper. 

\begin{table}[th!]
\centering
\small
 \renewcommand{\arraystretch}{1.1}
 \renewcommand{\tabcolsep}{1.4mm}
 \caption{Comparing the proposed approach with additional baselines from \cite{yu2019attributing,wang2020cnn}. * represents the pretrained features used for clustering while $\crosssymbol$ denotes the features obtained from binary classification, the original task of \cite{wang2020cnn}. We also provide a baseline (denoted by \#) using our approach but with JPEG and blur augmentations as used by \cite{wang2020cnn}. This does worse on both clustering metrics compared to our original approach.}
    \vspace{5pt}
\begin{tabular}{@{}llcc@{}}
\toprule
                     
\makecell{\textbf{\# of}\\ \textbf{clusters}}& 
\textbf{Method}& \multicolumn{1}{c}{\textbf{Avg. Purity}} & \multicolumn{1}{c}{\textbf{NMI}} \\ 
\midrule
\multirow{3}{*}{20}& Wang \etal \cite{wang2020cnn}*& 0.1946& 0.2042  \\   &Yu \etal \cite{yu2019attributing}*& 0.4529& 0.4543\\        
&Wang \etal \cite{wang2020cnn}$\crosssymbol$&0.3841&0.4434\\ \hdashline 
\multirow{3}{*}{500}& Wang \etal \cite{wang2020cnn}*& 0.2929& 0.2004  \\   &Yu \etal \cite{yu2019attributing}*& 0.5947& 0.3916\\        
&Wang \etal \cite{wang2020cnn}$\crosssymbol$&0.6082&0.4334\\\hdashline
258&$\textnormal{Ours}^\#$& 0.7696& 0.6249\\
266&Ours&0.8216& 0.6552\\ \bottomrule
\end{tabular}
\label{tab:additional_baselines}
\end{table}

\begin{table*}[]
    \centering
    \small
    \renewcommand{\arraystretch}{1.0}
    \renewcommand{\tabcolsep}{6pt}
    \caption{Comparison of our approach using WTA hash with ODIN \cite{liang2017enhancing}. The in-distribution dataset is CIFAR-100 which is used to train a DenseNet. We evaluate our method on the same metrics reported in \cite{liang2017enhancing}. The numbers reported are in the format of "ODIN/Ours". All values are in percentages. $\uparrow$ implies that the larger value is better while $\downarrow$ implies smaller value is better. We outperform ODIN on all OOD datasets and metrics excluding LSUN (crop).}
    \vspace{5pt}
    \begin{tabular}{@{}lccccc@{}}
    \toprule
    Out-distribution dataset & FPR at 95\% TPR $\downarrow$& Detection error $\downarrow$ & AUROC $\uparrow$& AUPR In $\uparrow$& AUPR Out$\uparrow$\\
    \midrule
    Tiny-ImageNet (crop)&$26.9/\boldsymbol{18.8}$&$12.9/\boldsymbol{10.2}$&$94.5/\boldsymbol{96.4}$&$94.7/\boldsymbol{96.6}$&$94.5/\boldsymbol{96.3}$\\
    Tiny-ImageNet (resize)&$57.0/\boldsymbol{20.2}$&$22.7/\boldsymbol{10.6}$&$85.5/\boldsymbol{96.2}$&$86.0/\boldsymbol{96.3}$&$84.8/\boldsymbol{96.1}$\\
    LSUN (crop)&$\boldsymbol{18.6}/32.1$&$\boldsymbol{9.7}/14.0$&$\boldsymbol{96.6}/93.8$&$\boldsymbol{96.8}/94.2$&$\boldsymbol{96.5}/93.7$\\
    LSUN (resize)&$58.0/\boldsymbol{17.3}$&$22.3/\boldsymbol{9.6}$&$86.0/\boldsymbol{96.8}$&$87.1/\boldsymbol{97.0}$&$84.8/\boldsymbol{96.7}$\\
    iSUN&$64.9/\boldsymbol{28.3}$&$24.0/\boldsymbol{12.6}$&$84.0/\boldsymbol{94.8}$&$85.1/\boldsymbol{95.2}$&$81.8/\boldsymbol{94.6}$\\
    Gaussian&$100.0/\boldsymbol{0.0}$&$17.9/\boldsymbol{0.1}$&$99.5/\boldsymbol{100.0}$&$87.5/\boldsymbol{100.0}$&$65.1/\boldsymbol{99.8}$\\
    Uniform&$100.0/\boldsymbol{0.0}$&$38.0/\boldsymbol{0.0}$&$40.5/\boldsymbol{100.0}$&$60.5/\boldsymbol{100.0}$&$40.9/\boldsymbol{99.9}$\\
    
    \bottomrule
    \end{tabular}
    \vspace{-0.1in}
    \label{tab:ood_related}
\end{table*}
\section{Out-of-distribution detection}
In this section, we provide more details on the WTA hash and also a comparison between cosine based distance and the WTA hashing based hamming distance for out-of-distribution detection. Additionally, we compare our approach with another popular out-of-distribution algorithm \cite{liang2017enhancing} and show that our algorithm performs well on their reported benchmarks. Finally, we analyze the effect of the percentile threshold used in our approach.
\subsection{WTA hash details}
The WTA hashing algorithm proceeds as follows. Suppose a single feature vector $\boldsymbol{x}$ has a dimension $d$. We generate $H$ different permutations $\boldsymbol{p}_i, \ i\in\{1,...,H\}$ of indices $\{1,...,d\}$ and then apply each of these permutations to $\boldsymbol{x}$ to get a set of vectors $\{\boldsymbol{x}'_i\}_{i=1}^{H}$. For each vector $\boldsymbol{x}'_i$, we take the first $K$ elements, for a window size $K$, and obtain the index of the max element. The set of these $H$ indices (one for each permutation) yields a new vector $\boldsymbol{x}_H$. Note that $\boldsymbol{x}_H$ is a $H$ dimensional vector with its elements taking integral values in $[0,K-1]$. The distance between two feature vectors is then defined as the hamming distance between their corresponding hashes.\\
\subsection{Cosine based distance details}
We compare the in-distribution, out-distribution and overall accuracy of our algorithm for the 12 seen classes (as described in Table 1 of the paper) using the WTA hash distance and a cosine-based distance. The results are shown in Table ~\ref{tab:ood_wta_cosine}. As the number of samples in our in-distribution is roughly the same as number of samples in our out-distribution, we use standard accuracy as our metrics for comparison. In-distribution accuracy refers to the accuracy on all the samples in the discovery set which belong to the 12 seen classes while out-distribution corresponds to those belonging to the 8 unseen classes. Net accuracy is the overall accuracy on the full discovery set. We see that using the hash outperforms cosine based distance in terms of the net accuracy and in-distribution accuracy. It performs lower than the cosine-based distance in terms of the out-distribution accuracy but with only a small difference. This is because of an inherent tradeoff between in-distribution and out-distribution accuracy based on the percentile threshold.
\begin{table}
    \centering
    \small
    \renewcommand{\arraystretch}{1.0}
    \renewcommand{\tabcolsep}{6pt}
    \caption{Comparison of our OOD step using WTA hash or cosine distance. We see that the WTA hash consistently outperforms the cosine-based distance at all 4 iterations of training even though it drops slightly on the out-distribution accuracy.}
    \vspace{5pt}
    \begin{tabular}{@{}cccc@{}}
    \toprule
    Iteration& \makecell{In-distribution\\ Accuracy (\%)} & \makecell{Out-distribution\\ Accuracy (\%)}&\makecell{Net\\ Accuracy (\%)}\\
    \midrule
    1& $86.02/\boldsymbol{91.74}$&$\boldsymbol{93.26}/89.35$&$89.33/\boldsymbol{90.65}$\\
    2& $83.49/\boldsymbol{88.33}$&$\boldsymbol{98.36}/97.63$&$90.22/\boldsymbol{92.58}$\\
    3& $81.14/\boldsymbol{85.37}$&$\boldsymbol{99.32}/98.34$&$89.45/\boldsymbol{91.3}$\\
    4& $79.11/\boldsymbol{82.94}$&$99.10/\boldsymbol{99.12}$&$88.27/\boldsymbol{90.33}$\\
    \bottomrule
    \end{tabular}
    \vspace{-0.1in}
    \label{tab:ood_wta_cosine}
\end{table}

\begin{table}
    \centering
    \footnotesize
    \renewcommand{\arraystretch}{1.1}
    \renewcommand{\tabcolsep}{1.4mm}
    \caption{Comparison between using the WTA hash based hamming distance or the cosine based distance for computing the 1-NN graph during merge step. We analyze the performance directly at iteration 1 and also at the end of 4 iterations for both stages of merge and refine.}
    \label{tab:merge_wta_cosine}
    \vspace{5pt}
    \resizebox{\linewidth}{!}{
    \begin{tabular}{@{}ccccccc@{}}
    \toprule
    \#iter. & Stage&\makecell{Avg.\\ Purity}& NMI &\makecell{\% Samples\\ Discovered} &\makecell{ \# of Sources\\ Discovered} & \makecell{\# of\\clusters}\\
    \midrule
    \multirow{2}{*}{1} &Merge& $0.791/\boldsymbol{0.793}$& $0.642/\boldsymbol{0.646}$&-&-&$433/\boldsymbol{383}$\\
    &Refine& $0.776/\boldsymbol{0.780}$& $\boldsymbol{0.671}/0.666$&$74.70/\boldsymbol{76.54}$&$4/\boldsymbol{5}$&$\boldsymbol{167}/180$\\
    \multirow{2}{*}{4}&Merge& $0.820/\boldsymbol{0.825}$& $0.635/\boldsymbol{0.647}$&-&-&$618/\boldsymbol{432}$\\
    &Refine& $0.819/\boldsymbol{0.823}$& $0.651/\boldsymbol{0.655}$&$91.80/\boldsymbol{94.76}$&$\boldsymbol{8}/\boldsymbol{8}$&$294/\boldsymbol{266}$\\
    \bottomrule\\
    \end{tabular}
    }
\small
\caption{Reducing number of clusters for K-Means (K) by almost half at each iteration. Average Purity and NMI does not change drastically compared to our default setup.}
\vspace{5pt}
\label{tab:red_k}
\resizebox{0.6\linewidth}{!}{
    \begin{tabular}{cccc}
    \toprule
    Iteration & K & Avg. Purity & NMI\\
    \midrule
    1& 500 & 0.695 & \textbf{0.6756} \\
2&250 & 0.7877 & 0.6572\\
3&125 & \textbf{0.8086} & 0.6484\\
4&60 & 0.8056 & 0.6399\\
\multicolumn{2}{c}{Ours (Default)}&0.8216&0.6552\\
    \bottomrule\\
    \end{tabular}
}
\caption{We Na\"ively recluster the test set after each training step and use them as pseudolabels for retraining. Compared to our original approach, a significant drop in Average Purity and NMI is observed.}
\vspace{5pt}
\label{tab:recluster}
\resizebox{0.8\linewidth}{!}{
\begin{tabular}{ccccc}
\toprule
\multirow{2}{*}{Step} & \multicolumn{2}{c}{Avg. Purity} & \multicolumn{2}{c}{NMI}\\
& Reclustering & Ours & Reclustering & Ours\\
\midrule
1  & 0.7858 & 0.7803   & 0.5402 & 0.6658 \\
2  & 0.7803 & 0.8183    & 0.5383 & 0.6625 \\
3  & 0.7597 & 0.8211    & 0.5289 & 0.6595   \\
4  & 0.7303 & 0.8216  & 0.5112 & 0.6552 \\              \bottomrule
\end{tabular}}
\label{tab:recluster}
\end{table}

\subsection{Related works comparison for OOD}
We now compare our approach with the popular out-of-distribution (OOD) approach called ODIN \cite{liang2017enhancing} on their benchmark. We show results using features extracted from DenseNet trained on CIFAR-100. We evaluate on the various out-of-distribution datasets provided by the authors of \cite{liang2017enhancing} and report our results in Table \ref{tab:ood_related}. We see that we outperform their algorithm on almost all datasets except LSUN (crop). Additionally, our algorithm has very few hyperparameters which require careful tuning and does not require a validation set. This shows that our approach generalizes well to other dataset setups and can be used in general for the problem of out-of-distribution detection.

\subsection{Threshold analysis}
We evaluate the performance of our pipeline when varying the percentile threshold which was set by default to $0.9$. We run our full pipeline iteratively for $4$ iterations and report the numbers for $4$ different values of $0.7$, $0.8$, $0.9$, $0.95$. The results are summarized in Table \ref{tab:ood_thresh}. Increasing the threshold has the expected result of decreased clusters as more samples in the discovery set are called in-distribution and are attributed to existing clusters. However, the resulting cluster metrics do not vary drastically in the neighbourhood of the default value of $0.9$. This shows that our pipeline is fairly robust to the percentile threshold which also intuitively transfers across datasets. Our approach thus has a single easy to tune scalar which automatically varies the threshold for the different seen classes or clusters.
\begin{table}
    \centering
    \small
    \renewcommand{\arraystretch}{1.0}
    \renewcommand{\tabcolsep}{5pt}
    \caption{Effect of the percentile threshold for OOD on the final performance. The default value for our experiments is 0.9. For all the thresholds, all sources were discovered.}
    \vspace{5pt}
    \begin{tabular}{@{}ccccc@{}}
    \toprule
    $\beta$ & Avg. Purity & NMI & \makecell[c]{\# of \\Clusters} & \makecell[c]{\% Samples \\Discovered}\\
    \midrule
    0.7 &0.7952 &0.5842 &449 &0.9226\\
    0.8 & 0.8087 & 0.6135 & 376 &0.9234\\
    0.9 & 0.8216 & 0.6552 & 266 & 0.9476\\
    0.95 & 0.8268 & 0.6985 & 181 & 0.9358\\
    
    \bottomrule
    \end{tabular}
    \vspace{-0.1in}
    \label{tab:ood_thresh}
\end{table}

\section{Clustering}
We analyze the effect of the clustering, merge and refine stages of the pipeline, varying a number of hyperparameters and show our pipeline's robustness to these values with respect to the final performance. 
\subsection{Effect of different number of clusters and number of rounds of merge and refine}
The $k$ value chosen for K-Means changes the number of clusters that are passed on to the merge step.  Additionally, we try performing multiple rounds of merge and refine within a single iteration, which decreases the number of clusters and improve purity. Table ~\ref{tab:k_ai} summarizes the effect of changing $k$ and the number of rounds of merge and refine, $r$, for the pipeline. We see that for a fixed $r$ and different $k$'s, even though the number of clusters changes in the clustering stage, the final number of clusters at the end of $4$ iterations are similar and the performance on Average Purity and NMI does not vary drastically. However, when we fix $k$ and vary $r$ we see that final number of clusters show a visible drop and as expected NMI increases slightly while Average Purity decreases.

\begin{figure*}[t]
\begin{center}
  \includegraphics[width=\linewidth]{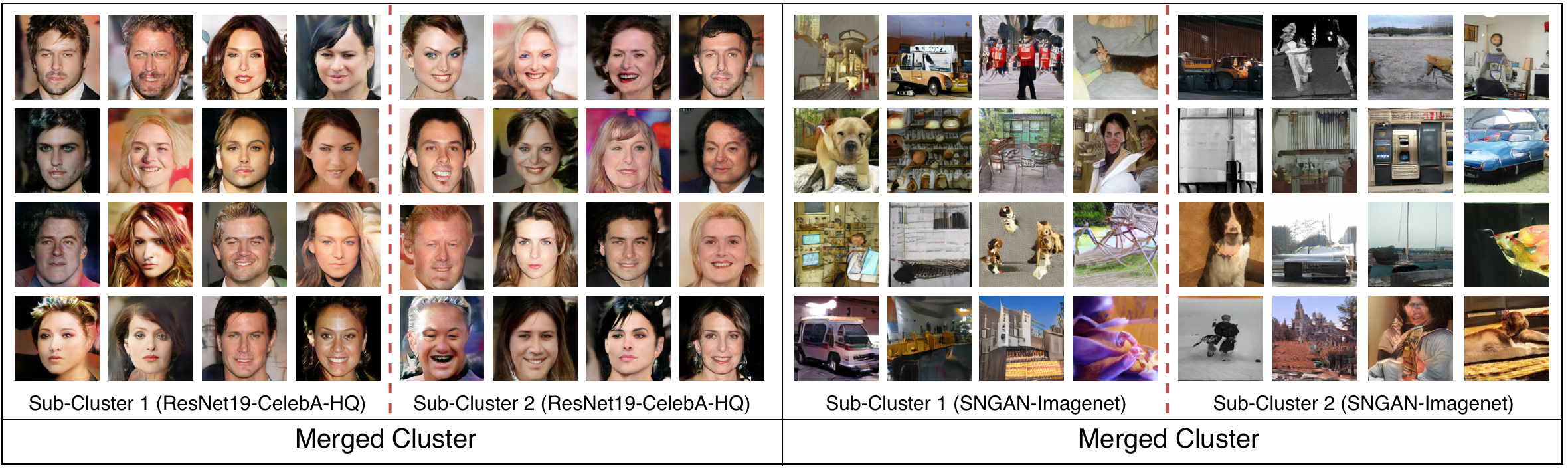}
\end{center}
  \caption{Some example clusters merged by our approach during the merge step.}
\label{fig:merge_vis}
\end{figure*}
\subsection{Cosine distance based merging}
Instead of using the hamming distance between the WTA hashes of the feature vectors, we use a cosine based distance between feature vectors and compare the performance at the end of the first iteration and also at the end of 4 iterations. Table \ref{tab:merge_wta_cosine} compares cosine and WTA based distance at both points in the pipeline at the end of both merge and refine steps. We see that the WTA hash based distance marginally outperforms the cosine based distance at almost all points in the pipeline. It also discovers a higher percentage of samples with fewer clusters demonstrating its effectiveness in our pipeline.

\subsection{Size threshold and SVM firing threshold}
Next, we evaluate the effect of varying the size threshold, $\tau$ for discarding clusters at the end of refine step, as well as varying the SVM fire threshold, $\epsilon$, which also controls the number of clusters (likely impure) being discarded. The results are summarized in Table ~\ref{tab:svm_thresh}. As expected increasing $\tau$ or $\epsilon$ decreases the number of clusters as more samples are discarded. This comes at the cost of fewer samples and GANs being discovered. However, the clustering metrics do not vary drastically showing that our pipeline is robust to these hyperparameter values and can generalize across varying datasets and sizes.

\begin{table}[]
\centering
\small
\renewcommand{\arraystretch}{1.1}
\renewcommand{\tabcolsep}{6pt}
\caption{Effect of varying the size threshold ($\tau$) and SVM fire fire threshold ($\epsilon$) on performance. The default setting corresponds to $\tau=100$ and $ \epsilon=0.5$.}
\vspace{5pt}
\resizebox{\linewidth}{!}{
\begin{tabular}{@{}ccccccc@{}}
\toprule
$\tau$ & \multicolumn{1}{c}{$\epsilon$} & \makecell[c]{Avg.\\ Purity} & \multicolumn{1}{c}{NMI} & \makecell[c]{Sources\\ Discovered} & \makecell[c]{\#\\ of Clusters} & \makecell[c]{\% Samples\\ Discovered} \\ \midrule
\multirow{3}{*}{50}                                         & 0.3                        & 0.8178   & 0.6356 & 20/20    & 407   & 0.9695 \\ 
& 0.5                                        & 0.8212                                    & 0.6326                            & 20/20                       & 434                                    & 0.9729                                              \\ 
& 0.7                                        & 0.8223                                    & 0.6551                            & 20/20                       & 308                                    & 0.9549                                              \\ \hdashline
\multirow{3}{*}{100} & 0.3  & 0.8238                                    & 0.6451                            & 20/20                       & 293                                    & 0.9590                                              \\ 
& 0.5                                        & 0.8216                                    & 0.6552                            & 20/20                       & 266                                    & 0.9476                                              \\ 
& 0.7                                        & 0.823                                     & 0.6573                            & 20/20                       & 245                                    & 0.9237                                              \\  \hdashline
\multirow{3}{*}{200}                                           & 0.3                    & 0.8265                                    & 0.6731                            & 20/20                       & 166                                    & 0.8919                                              \\ 
& 0.5                                        & 0.8197                                    & 0.67                              & 20/20                       & 161                                    & 0.9015                                              \\ 
& 0.7                                        & 0.8233                                    & 0.686                             & 19/20                       & 138                                    & 0.8841                                              \\  \hdashline
\multirow{3}{*}{300}                                           & 0.3                    & 0.7849                                    & 0.7124                            & 14/20                       & 70                                     & 0.8643                                              \\ 
& 0.5                                        & 0.8009                                    & 0.6992                            & 13/20                       & 89                                     & 0.8340                                              \\ 
& 0.7                                        & 0.8055                                    & 0.7161                            & 18/20                       & 85                                     & 0.8638                                              \\ 
\bottomrule
\end{tabular}
}
\label{tab:svm_thresh}
\end{table}

\begin{table}[]
\centering
\small
\renewcommand{\arraystretch}{1.1}
\renewcommand{\tabcolsep}{4pt}
\caption{Effect of varying the number of clusters $k$ for K-Means along with the number of times merge and refine (Additional Iters) is performed for each step. In the default setting Additional Iters is $0$ as merge and refine are performed once per step while $k=500$. }
\vspace{5pt}
\resizebox{\linewidth}{!}{
\begin{tabular}{@{}ccccccc@{}}
\toprule
K & \makecell[c]{Additional\\ Iters} & \makecell[c]{Avg.\\ Purity} & NMI & \makecell[c]{Sources\\ Discovered} & \makecell[c]{\% Samples\\ Discovered} & \makecell[c]{\# of\\ Clusters} \\ \midrule
\multirow{4}{*}{100} & 0  & 0.7948 & 0.6722 & 20/20  & 0.9839 & 133 \\
 & 1  & 0.7952 & 0.6938 & 20/20  & 0.9876 & 104\\
& 2    & 0.7889   & 0.7016 & 19/20 & 0.9904 & 89  \\
& 3   & 0.7627 & 0.7255  & 19/20  & 0.9835 & 66  \\ \hdashline
\multirow{4}{*}{200}  & 0 & 0.8086  & 0.657 & 20/20 & 0.9762 & 199 \\
& 1   & 0.8125 & 0.6798 & 20/20 & 0.9847 & 154\\
& 2 & 0.7944   & 0.6981   & 19/20  & 0.9749  & 105 \\
& 3   & 0.7944  & 0.7085 & 20/20 & 0.9757 & 95  \\ \hdashline
\multirow{4}{*}{300}  & 0   & 0.8142 & 0.652  & 20/20  & 0.9665 & 238 \\
& 1  & 0.8051 & 0.6696 & 20/20  & 0.9486  & 163  \\
& 2  & 0.802 & 0.6975  & 20/20  & 0.9602  & 121   \\
& 3 & 0.7669  & 0.7046   & 17/20                       & 0.9569 & 92  \\ \hdashline
\multirow{4}{*}{400}  & 0 & 0.8175  & 0.6472  & 20/20 & 0.9494    & 281\\
& 1  & 0.814   & 0.6639   & 20/20 & 0.9637   & 196 \\
& 2   & 0.8105   & 0.6817  & 20/20  & 0.9687 & 158 \\
& 3  & 0.8005 & 0.7078  & 20/20    & 0.9679  & 113 \\ \hdashline
\multirow{4}{*}{500} & 0   & 0.8216   & 0.6552   & 20/20    & 0.9476 & 266 \\
& 1  & 0.8249  & 0.6736 & 20/20  & 0.9532 & 195  \\
& 2  & 0.8118 & 0.6887  & 20/20 & 0.9536    & 136  \\
& 3  & 0.7862  & 0.691 & 20/20   & 0.9485 & 114\\ \hdashline
\multirow{4}{*}{600} & 0  & 0.8264 & 0.6535  & 20/20 & 0.9396  & 279\\
& 1  & 0.8229  & 0.667 & 20/20  & 0.9545     & 212 \\
& 2  & 0.8176  & 0.6854  & 20/20 & 0.9470   & 153\\
& 3  & 0.8039  & 0.6936  & 20/20  & 0.9459  & 121  \\ 
\hdashline
\multirow{4}{*}{700}  & 0 & 0.8279  & 0.6572  & 20/20  & 0.9137 & 266 \\
& 1 & 0.8294  & 0.6837 & 20/20  & 0.9115  & 172 \\
& 2  & 0.8348 & 0.6887 & 20/20 &0.9473  & 166  \\
& 3 & 0.8233 & 0.6923  & 20/20  & 0.9382 & 139  \\              
\bottomrule
\end{tabular}
}
\label{tab:k_ai}
\end{table}

\subsection{Reclustering}
We now evaluate the effect of removing the merge and refine steps from our pipeline. Our pipeline consists of the usual network training followed by clustering. For each iteration, we discard the old clusters and perform clustering again using K-Means on all the test set samples. We then use the new clusters as pseudo-labels and retrain our network for improving the features. It should be noted there is no OOD detection, merge or refine steps. This method is similar to ClusterFit \cite{yan2020clusterfit} who also use cluster pseudo-labels for network training. We analyze the results in Table~\ref{tab:recluster} by comparing with our original pipeline. We see that the performance drops by a significant margin showing that it is crucial to maintain existing clusters and iteratively merge and refine them while simultaneously improving the feature representations of the existing clusters.

\subsection{Effect of varying number of clusters}
For most of our experiments we run the clustering using K-Means at a fixed value of $k$, which is used for all iterations. As the number of undiscovered samples reduce as number of iterations increases, we evaluate our pipeline's performance by decreasing $k$ after each iteration. We thus, approximately halve the value of $k$ after each iteration. The results are reported in Table ~\ref{tab:red_k}. We see that the performance does not change drastically compared to the default setup which shows that our network does not heavily rely on the number of clusters used during K-Means. 

\subsection{Qualitative Analysis}
We now qualitatively show the effect of our merge step for a few clusters. The merge step of our approach merges clusters belonging to the same class but are actually fragmented due to overclustering from K-Means. We visualize two such clusters in Fig.~\ref{fig:merge_vis}. As we showed in the main paper, our clusters focus on the GAN source rather than image semantics and the merge step successfully combines clusters having the same majority GAN source.

\section{Network Training}
\subsection{Effect of faster training}
By default, we retrain all feature extractor weights every iteration. To reduce the cost of full network retraining, we analyze finetuning only the final residual block of the ResNet-50 backbone along with the subsequent fully connected layers of the feature extractor. We also analyze using a lighter network such as MobileNet \cite{howard2017mobilenets} for our network training and compare it with our original setup. Additionally, we try to see the effect of removing the merge step from the pipeline The results are summarized in Table \ref{tab:fast_training}. We see that there is a small drop in network performance in terms of both Average Purity and NMI and it also fails to discover a single unseen source. Constraining the network is likely to have restricted the network's capability of improving the discovery set features although it doesn't have a significant impact. On the other hand, MobileNet obtains a higher NMI because of much fewer clusters, albeit at the cost of not discovering most of the unseen sources. This shows that very light networks are not as effective in obtaining discriminative representations for discovering new sources. Also the performance without merge is sub-par to our original approach which shows the importance of performing merge step to group similar clusters together after over-clustering.

\begin{table}
    \centering
    \footnotesize
    \renewcommand{\arraystretch}{1}
    \renewcommand{\tabcolsep}{2pt}
    \caption{Results on our setup with slight variations in our training.}
    \begin{tabular}{@{}lcccc@{}}
    \toprule
    Experiment & \makecell[c]{\# of Clusters} & Avg. Purity & NMI & \makecell[c]{\# Sources Disc.}\\
    \midrule
    Ours (Original) & 209 & 0.861 & 0.724 & 20/20\\
    Ours (w/o merge)& 257 & 0.841 & 0.712 & 19/20\\
    Ours (Freeze) & 229 & 0.850 & 0.691 & 19/20\\
    MobileNet & 70 & 0.846 & 0.773 & 15/20\\
    \bottomrule
    \end{tabular}
    \label{tab:fast_training}
\end{table}

\subsection{Effect of image size}
By default, we resize and center crop all images to $256\times256$ in most of our experiments. We compare results when image sizes are varied from $64$ to $256$ for network training. From Table \ref{tab:image_size}, we see that increasing image size shows a marked improvement in all metrics. Therefore, we hypothesize that model fingerprints are likely more detectable and distinguishable when the image is resized to a higher resolution. However, this comes at the cost of increasing network training times and memory requirements (quadratically) which is infeasible in an online setup or for very large scale datasets.

\begin{table}
    \centering
    \footnotesize
    \renewcommand{\arraystretch}{1}
    \renewcommand{\tabcolsep}{2pt}
    \caption{Results on our setup with varying image sizes.}
    \begin{tabular}{@{}lcccc@{}}
    \toprule
    Image Size & \makecell[c]{\# of Clusters} & Avg. Purity & NMI & \makecell[c]{\# Sources Disc.}\\
    \midrule
    64 & 169 & 0.655 & 0.579 & 19/20\\
    128 & 266 & 0.822 & 0.673 & 20/20 \\
    256 & 209 & 0.861 & 0.724 & 20/20\\
    \bottomrule
    \end{tabular}
    \label{tab:image_size}
\end{table}

\section{Multiple seed sources}
\cite{yu2019attributing} showed that training generators with different random seeds generate different distinguishable fingerprints in their images. We analyze whether we can discover new separate sources when a single generator architecture is trained on the same dataset but with different seeds. Table \ref{tab:multiple_seeds} shows results comparing this setup with our original setup. We add 2 different seeds for ProGAN for both CelebA and LSUN-Bedroom providing 2 new sources. Note that only a single seed of ProGAN trained on LSUN-Bedroom is present in the labeled set while the other 3 sources are unseen. The remaining classes are same as our original setup as described in Table 1 of the main paper. We see that there is only a small drop in Average Purity and NMI although it fails to discover a single unseen source.
\begin{table}
    \centering
    \footnotesize
    \renewcommand{\arraystretch}{1}
    \renewcommand{\tabcolsep}{2pt}
    \caption{Results on our setup with variations in training.}
    \begin{tabular}{@{}lcccc@{}}
    \toprule
    Experiment & \makecell[c]{\# of Clusters} & Avg. Purity & NMI & \makecell[c]{\# Sources Disc.}\\
    \midrule
    Ours (Original) & 209 & 0.861 & 0.724 & 20/20\\
    Ours + Unseen seeds & 216 & 0.842 & 0.702 & 21/22\\
    \bottomrule
    \end{tabular}
    \label{tab:multiple_seeds}
\end{table}



\end{document}